\documentclass[conference]{IEEEtran}
\IEEEoverridecommandlockouts

\usepackage{cite}
\usepackage{amsmath,amssymb,amsfonts}
\usepackage{algorithm}
\usepackage{algpseudocode}
\usepackage{graphicx}
\usepackage{booktabs}

\usepackage{subcaption}
\usepackage{textcomp}
\usepackage{xcolor}
\def\BibTeX{{\rm B\kern-.05em{\sc i\kern-.025em b}\kern-.08em
    T\kern-.1667em\lower.7ex\hbox{E}\kern-.125emX}}
\begin{document}

\newcommand{\todo}[1]{\textbf{[TODO: #1]}}
\definecolor{inteins}{RGB}{128,179,255}
\newcommand{\TODO}[2]{\textcolor{blue}{TODO: \textbf{\red{#1}} #2}}
\newcommand{\CHECK}[1]{\textcolor{red}{#1}\textcolor{blue}{ $<$--CHECK}}
\newcommand{\REV}[1]{\textcolor{blue}{#1}}

\newcommand\red[1]{{\color{red}#1}} 
\newcommand\blue[1]{{\color{blue}#1}} 

\newcommand{\calL}{\mathcal{L}}
\newcommand{\calG}{\mathcal{G}}
\newcommand{\calD}{\mathcal{D}}
\newcommand{\calU}{\mathcal{U}}
\newcommand{\calV}{\mathcal{V}}
\newcommand{\calS}{\mathcal{S}}
\newcommand{\calR}{\mathbb{R}}
\newcommand{\calE}{\mathbb{E}}
\newcommand{\MNIST}{\texttt{MNIST}\xspace}
\newcommand{\RING}{\texttt{RING}\xspace}
\newcommand{\BLOB}{\texttt{BLOB}\xspace}

\newcommand{\methodLong}{CO-evolutionary Multi-Objective Discriminator SSL-GAN}
\newcommand{\method}{COMOD-SSLGAN}

\title{Population-Based Multi-Objective Training of Discriminators for Semi-Supervised GANs}

\author{
\IEEEauthorblockN{Francisco Sedeño}
\IEEEauthorblockA{
\textit{ITIS Software},
\textit{University of Malaga} \\
Malaga, Spain \\
0009-0000-7797-3562
}
\and
\IEEEauthorblockN{Francisco Chicano}
\IEEEauthorblockA{
\textit{ITIS Software},
\textit{University of Malaga} \\
Malaga, Spain \\
0000-0003-1259-2990
}
\and
\IEEEauthorblockN{Jamal Toutouh}
\IEEEauthorblockA{
\textit{ITIS Software},
\textit{University of Malaga} \\
Malaga, Spain \\
\textit{CSAIL MIT}, Cambridge, MA, USA\\
0000-0003-1152-0346
}

}

\maketitle

\begin{abstract}
Semi-supervised generative adversarial networks (SSL-GANs) can exploit large unlabeled datasets while retaining a classifier in the discriminator, but their training is often unstable. This paper proposes a population-based evolutionary training strategy in which discriminator learning is formulated as a multi-objective optimization problem. Instead of aggregating the supervised and unsupervised components of the SSL objective into a single scalar loss, the method maintains a population of discriminators ranked by Pareto dominance, enabling the exploration of different trade-offs between classification accuracy and real/fake discrimination. This formulation aims to improve both roles of SSL-GANs: learning accurate classifiers and training generators capable of producing realistic samples. We analyze several variants, including an elitist strategy and a mono-objective ablation, to assess the role of multi-objective selection. Experiments on MNIST with limited labels show improved training robustness compared to SSL-GAN and CE-SSL-GAN state-of-the-art baselines, while the elitist variant consistently achieves the highest classification accuracy.
\end{abstract}

\section{Introduction}
\label{sec:introduction}

A generative adversarial network (GAN)~\cite{goodfellow2020generative} consists of two neural networks trained in an adversarial setting: a \emph{generator}, which produces synthetic samples, and a \emph{discriminator}, which distinguishes real from generated data. The generator is rewarded when its samples are classified as real, while the discriminator is rewarded for correctly identifying real and fake samples. GANs can learn complex data distributions and have been successfully applied in domains such as medical image generation~\cite{DBLP:journals/asc/WangWY21,flores2022coevolutionary}.

Training GANs is challenging due to the non-stationary interaction between generator and discriminator, which may lead to issues such as vanishing gradients or mode collapse. Among different approaches to mitigate GAN training pathologies, evolutionary computation (EC) has been proposed as a mechanism to stabilize this process by maintaining multiple interacting models. For example, Lipizzaner combines competitive co-evolution with spatially distributed populations inspired by cellular evolutionary algorithms~\cite{hemberg2021spatial}.

GANs are also well suited for semi-supervised learning (SSL) to learn classifiers, where datasets contain both labeled and unlabeled samples. In SSL-GANs, the discriminator acts both as a classifier and as a real/fake detector. This framework has been widely studied~\cite{DBLP:conf/ACISicis/TachibanaMU16}, with several surveys summarizing recent developments~\cite{app12031718,Ma2024}.

In literature, the application of co-evolutionary algorithms to SSL-GANs is recent. Previous works explored spatial co-evolutionary populations to improve classification and generation quality~\cite{toutouh2023semi,toutouh2023semiGECCO}. Recently, Sedeño et al.~\cite{sedeno2025} proposed Co-evolutionary Elitist SSL-GAN (CE-SSL-GAN), introducing panmictic populations, elitist replacement, and multiple offspring generation.
These approaches optimize the discriminator using a single scalar loss obtained by aggregating the supervised and unsupervised SSL objectives. However, these objectives correspond to different learning tasks: classifying labeled samples and distinguishing real from generated data.

In this work, we propose, \methodLong\ (\method), a population-based SSL-GAN training strategy in which discriminator learning is formulated as an explicit multi-objective optimization problem. Instead of aggregating the supervised and unsupervised losses, we treat them as separate objectives and apply Pareto-based selection to maintain a diverse population of discriminators representing different trade-offs between classification and adversarial discrimination.

Thus, the main contributions of this work are:
(a) a population-based SSL-GAN training framework where discriminator learning is formulated as a bi-objective optimization problem separating supervised and unsupervised losses;
(b) a Pareto-based discriminator selection mechanism that preserves diverse trade-offs during training; and
(c) an empirical analysis of evolutionary variants, including elitist replacement and a mono-objective ablation corresponding to CE-SSL-GAN~\cite{sedeno2025}.

The remainder is organized as follows. Section~\ref{sec:background} introduces background. Section~\ref{sec:method} describes the proposed method. Section~\ref{sec:experimental-setup} and Section~\ref{sec:results} present the experimental setup and the results. Section~\ref{sec:conclusions} concludes the paper.

\section{Background}
\label{sec:background}

Let us assume a partially labeled dataset containing samples belonging to $K$ classes. Each sample is represented as a $d$-dimensional real vector $x \in \mathbb{R}^{d}$. For labeled samples, the associated class is encoded as a one-hot vector $y \in \{0,1\}^{K}$.
We denote the labeled real samples as
$p_{la} \subseteq \mathbb{R}^{d} \times \{0,1\}^{K}$ and the unlabeled real samples as
$p_{un} \subseteq \mathbb{R}^{d}$.

\begin{figure}[!ht]
\centering
\includegraphics[width=0.99\linewidth]{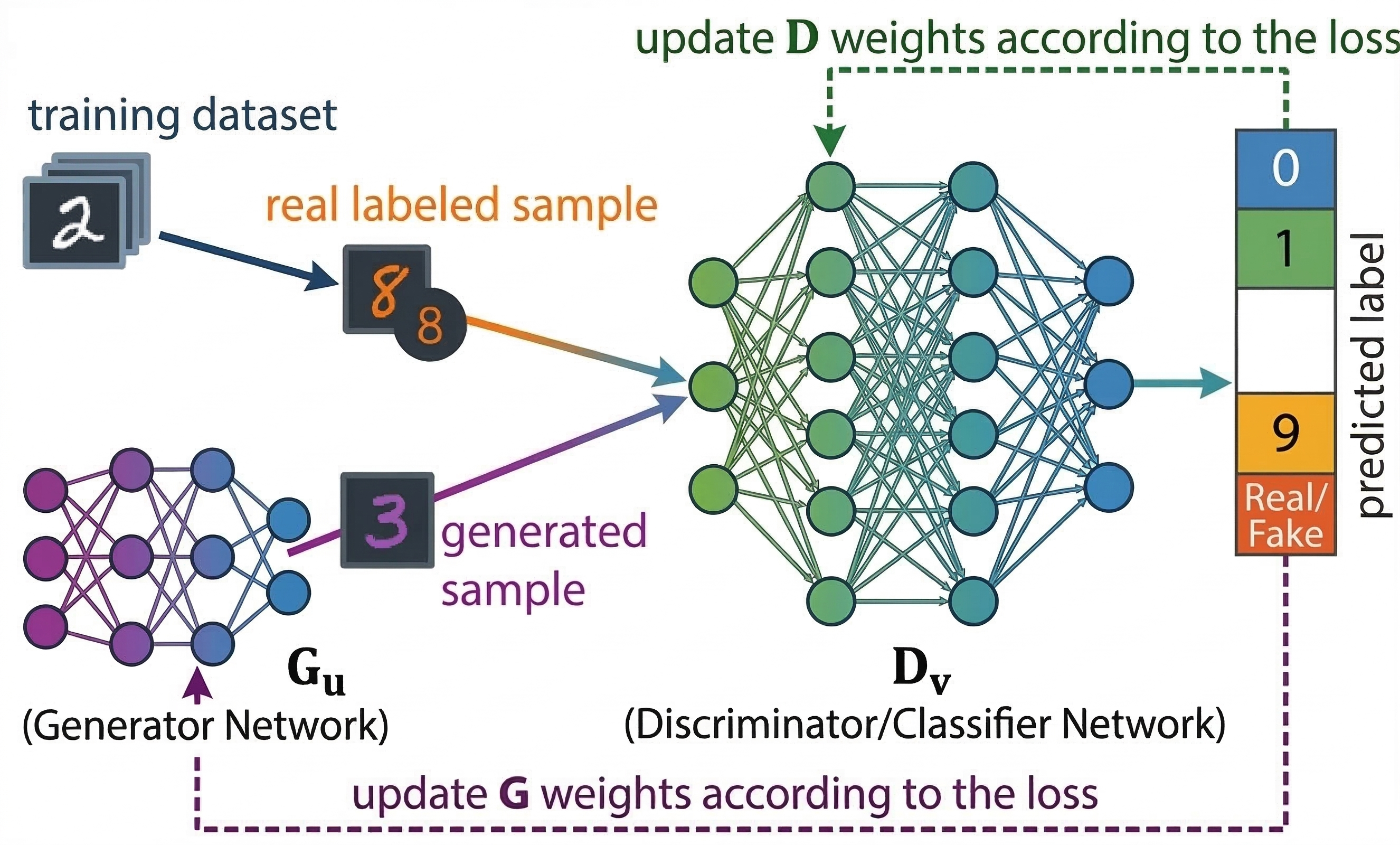}
\caption{SSL-GAN architecture.
The generator maps latent noise to synthetic samples, while the
discriminator outputs probabilities for the $K$ real classes and an
additional fake class.}
\label{fig:sslgan-representation}
\end{figure}

A SSL-GAN consists of two neural networks: a
generator and a discriminator (Fig.~\ref{fig:sslgan-representation}).
The generator $G_u$ maps a random vector from a latent space to the data space.
Formally, \hbox{$G_u : \mathbb{R}^{\ell} \rightarrow \mathbb{R}^{d}$}, where $\ell$ is the dimension of the latent space and $u$ denotes the
generator parameters.

The discriminator $D_v$ outputs a probability distribution over $K+1$
classes: the first $K$ correspond to the real classes, and the
$(K+1)$-th corresponds to the samples produced by the generator.
This \emph{$K+1$ trick} allows the discriminator to act simultaneously as a classifier and as a real/fake detector.
Formally, $D_v : \mathbb{R}^{d} \rightarrow [0,1]^{K+1}$, where $v$ denotes the discriminator parameters.
The first $K$ components represent the class probabilities,
denoted as $D_v^{class}(x) \in [0,1]^K$,
while the last component represents the probability that $x$ is generated,
denoted as $D_v^{fake}(x)$.

The objective of the generator is to produce samples that resemble the
distribution of real data.
Thus, the generator loss decreases when generated samples are classified
as real by the discriminator.
The generator loss is defined as \hbox{$\mathcal{L}_{G} =
\mathbb{E}_{z\sim \mathcal{N}^{\ell}(\mathbf{0},\mathbf{I})}
\left[
-\log\left(1 - D_v^{fake}(G_u(z))\right)
\right]$}.

The discriminator has two objectives:
(i) classify labeled samples into the correct class and
(ii) distinguish real samples from generated ones.
Thus, its loss is composed of two terms
$\mathcal{L}_{D} = \mathcal{L}_{D,s} + \mathcal{L}_{D,u}$,
where $\mathcal{L}_{D,s}$ is the supervised loss over labeled samples and
$\mathcal{L}_{D,u}$ is the unsupervised loss over unlabeled and generated
samples.

The supervised loss is

\begin{align}
\mathcal{L}_{D,s} =
\mathbb{E}_{(x,y)\sim p_{la}}
\left[
\sum_{i=1}^{K} y_i (-\log D_{v,i}^{class}(x))
\right],
\end{align}

while the unsupervised loss is

\begin{align}
\mathcal{L}_{D,u} &=
\mathbb{E}_{x\sim p_{un}}
\left[-\log(1-D_v^{fake}(x))\right] \\
&+
\mathbb{E}_{z\sim \mathcal{N}^{\ell}(\mathbf{0},\mathbf{I})}
\left[-\log(D_v^{fake}(G_u(z)))\right],
\end{align}

\noindent
where $D_{v,i}^{class}(x)$ denotes the $i$-th component of the output vector $D_v^{class}(x)$ in the discriminator. 

\subsection{SSL-GAN training}

SSL-GAN training alternates between updating the discriminator and the
generator.
The discriminator is trained using labeled and unlabeled real samples
together with generated samples, while the generator is trained to produce
samples that the discriminator classifies as real.

Algorithm~\ref{alg:sslgan-training} summarizes the standard SSL-GAN
training procedure.
The training proceeds over $T$ training epochs, iterating through batches of real samples (labeled, $B_L$, and unlabeled, $B_U$) and latent space vectors ($\mathbf{z}$) to compute $\mathcal{L}_{D,s}$, $\mathcal{L}_{D,u}$, $\mathcal{L}_{D}$, and $\mathcal{L}_{G}$.
The two loss functions, $\mathcal{L}_{D}$ and $\mathcal{L}_{G}$, are used to update the parameters of the discriminator and the generator, respectively, by stochastic gradient descent.

\begin{algorithm}[!h]
\small
\caption{SSL-GAN training}
\label{alg:sslgan-training}
\begin{algorithmic}[1]
\Require
$T$: training epochs,
$p_{la}$: labeled dataset,
$p_{un}$: unlabeled dataset,
$B_s$: batch size
\State Initialize generator $g$ and discriminator $d$
\For{$t=1$ to $T$}
    \For{batches $B_L \subseteq p_{la}$ and $B_U \subseteq p_{un}$}
    
        \State Sample latent noise $\mathbf{z}$
        \State $X_f \gets g(\mathbf{z})$ \Comment{Generate fake samples}
        
        \State Compute $\mathcal{L}_{D,s}$ using $B_L$
        \State Compute $\mathcal{L}_{D,u}$ using $B_U$ and $X_f$
        \State Update discriminator using $\mathcal{L}_{D,s}+\mathcal{L}_{D,u}$
        
        \State Sample new latent noise $\mathbf{z}$
        \State $X_f \gets g(\mathbf{z})$
        \State Compute $\mathcal{L}_{G}$
        \State Update generator
        
    \EndFor
\EndFor
\State \Return trained generator and discriminator
\end{algorithmic}
\end{algorithm}

\subsection{Related work}
\label{sec:related-work}

Training GANs is challenging due to the adversarial interaction between generator and discriminator.
This interaction creates a non-stationary optimization process that can lead to training pathologies such as oscillatory dynamics and mode collapse.
These difficulties become even more pronounced in semi-supervised settings, where the discriminator must simultaneously perform classification and real/fake discrimination.
Consequently, several works have explored alternative training strategies, including evolutionary and population-based approaches, to improve GAN training stability.

Evolutionary Computation (EC) has been widely investigated as a mechanism to improve GAN training.
An early example is Evolutionary GAN (EGAN)~\cite{wang2019evolutionary}, which maintains multiple generators trained with different loss functions and applies evolutionary selection to identify the best-performing models.
Similarly, Multi-objective Evolutionary GAN (MO-EGAN)~\cite{baioletti2020multi} addresses GAN training as a multi-objective optimization problem, balancing conflicting criteria such as fidelity and diversity.
Other approaches apply evolutionary search directly to GAN parameters, such as Differential Evolution GAN (DE-GAN)~\cite{zheng2019differential}, which uses differential evolution to explore the parameter space during training.

Coevolutionary algorithms, where generators and discriminators evolve in separate populations, have also been proposed.
For example, COEGAN~\cite{costa2019coegan,costa2020neuroevolution} coevolves generator and discriminator architectures that cooperate to produce high-quality synthetic samples.
More generally, competitive coevolution has proven effective in adversarial learning scenarios by promoting continuous adaptation between competing agents~\cite{hemberg2021spatial,toutouh2020parallel,toutouh2019spatial,toutouh2020data,toutouh2021signal,toutouh2023semi}.
Typically, individuals exchange components with neighboring models based on performance, creating distributed coevolutionary dynamics that help mitigate GAN training instabilities.
Casas and Toutouh~\cite{casas2025} investigates different coevolutionary replacement schemes, including $(\mu,\lambda)$ and $(\mu+\lambda)$ strategies, highlighting the importance of balancing exploration and exploitation in population-based adversarial learning.



Although many evolutionary GAN approaches focus on unsupervised generative modeling, some studies have explored their application to SSL.
The spatial coevolutionary SSL-GAN framework proposed in~\cite{toutouh2023semi,toutouh2023semiGECCO} introduces a distributed population of GANs arranged in a two-dimensional grid, where neighboring models exchange components based on performance.
This approach improves both classification accuracy and image generation quality when only a limited number of labeled samples is available. 
A recent approach is CE-SSL-GAN)~\cite{sedeno2025}, which introduces panmictic populations, elitist replacement, and multiple offspring generation per iteration, showing that generating multiple offspring can improve both classification accuracy and sample quality.

Despite these advances, most existing evolutionary SSL-GAN approaches rely on predefined evolutionary replacement schemes or spatial population structures.
In contrast, \method\ models discriminator training as a multi-objective optimization problem and applies Pareto-based selection to maintain a diverse set of discriminators during training, explicitly capturing the trade-off between supervised and unsupervised objectives within a population-based framework.

\section{\methodLong}
\label{sec:method}

This section presents \method, a co-evolutionary method to train SSL-GANs using small populations.
\method\ combines a a $(\mu + \lambda)$ competitive co-evolution with the standard SSL-GAN training loop described in Section~\ref{sec:background}. The key idea is to maintain \emph{two} populations (generators and discriminators) and to formulate discriminator training as a \emph{bi-objective} optimization problem, where supervised and unsupervised losses are treated as separate objectives. This enables the algorithm to preserve diverse discriminator behaviors along the trade-off between classification and discrimination.

This section describes the main components of \method.
We first define the generator and discriminator populations and formulate discriminator training as a bi-objective optimization problem.
We then present the competitive evaluation mechanism used to assess individuals through matchups between populations.
Next, we describe the variation operator, where stochastic gradient descent acts as a mutation operator by training coupled SSL-GANs.
Finally, we introduce the selection and elitism mechanisms and summarize the overall training procedure.

\subsection{Populations and objectives}
\label{subsec:method-populations}

The method operates over two populations of neural networks of the same size $\mu$:
a population of generators \hbox{$\mathbf{G}=\{G_{u_1},\ldots,G_{u_\mu}\}$} and a population of discriminators \hbox{$\mathbf{D}=\{D_{v_1},\ldots,D_{v_\mu}\}$}.
Each individual generator $G_{u_i}$ and discriminator $D_{v_j}$ follows the notation in Section~\ref{sec:background}, where $u_i$ and $v_j$ denote their parameters, respectively.

Given a discriminator $D_{v}$ and a generator $G_{u}$, we define the generator loss $\mathcal{L}_{G}(u;v)$ as in Section~\ref{sec:background}:
\begin{equation}
\mathcal{L}_{G}(u;v) =
\mathbb{E}_{z\sim \mathcal{N}^{\ell}(\mathbf{0},\mathbf{I})}
\left[
-\log\left(1 - D_v^{fake}(G_u(z))\right)
\right].
\end{equation}

The discriminator loss $\mathcal{L}_{D}(v;u)$ keeps the decomposition into supervised $\mathcal{L}_{D,s}(v)$ and unsupervised $\mathcal{L}_{D,u}(v;u)$ terms,
where $\mathcal{L}_{D,s}$ depends on labeled samples from $p_{la}$ and $\mathcal{L}_{D,u}$ depends on unlabeled samples from $p_{un}$ and fake samples produced by $G_u$ (see Section~\ref{sec:background}).
Instead of collapsing $\mathcal{L}_{D,s}$ and $\mathcal{L}_{D,u}$ into a single scalar, we treat discriminator learning as a bi-objective minimization problem:
\begin{equation}
\min_{v}\ \left(\mathcal{L}_{D,s}(v),\ \mathcal{L}_{D,u}(v;u)\right).
\label{eq:mo_discriminator}
\end{equation}

This formulation explicitly captures the two complementary roles of the discriminator in SSL-GANs: acting as a classifier for labeled data and distinguishing real from generated samples. By optimizing these objectives jointly through Pareto dominance and diversity preservation~\cite{deb2002fast}, the algorithm aims to maintain a set of discriminators representing different trade-offs between classification performance and adversarial discrimination. Ultimately, this encourages the training process to produce SSL-GAN models that simultaneously learn strong discriminative classifiers and effective generative models.

\subsection{Competitive evaluation and fitness aggregation}
\label{subsec:method-evaluation}

Because the loss of each discriminator depends on the current generators (and vice versa), individual quality is assessed through \emph{competitive matchups} between both populations.
Let $\mathcal{M}\subseteq\{1,\ldots,\mu\}\times\{1,\ldots,\mu\}$ be a set of matchups between discriminators and generators.
For each matchup $(i,j)\in\mathcal{M}$, we compute $\mathcal{L}_{D,s}(v_i)$, $\mathcal{L}_{D,u}(v_i;u_j)$, and $\mathcal{L}_{G}(u_j;v_i)$ using mini-batches drawn from $p_{la}$ and $p_{un}$ (and latent noise).
To obtain a population-level evaluation, we aggregate objectives across opponents. Concretely, we define the aggregated discriminator objectives as
\begin{align}
\widehat{\mathcal{L}}_{D,s}(v_i) &= \mathcal{L}_{D,s}(v_i), \\
\widehat{\mathcal{L}}_{D,u}(v_i) &= \frac{1}{|\mathcal{O}(i)|}\sum_{j\in \mathcal{O}(i)} \mathcal{L}_{D,u}(v_i;u_j),
\end{align}
where $\mathcal{O}(i)=\{j\mid (i,j)\in\mathcal{M}\}$ is the set of generators faced by discriminator $D_{v_i}$.
Similarly, the aggregated generator loss is
\begin{equation}
\widehat{\mathcal{L}}_{G}(u_j)=\frac{1}{|\mathcal{O}(j)|}\sum_{i\in\mathcal{O}(j)} \mathcal{L}_{G}(u_j;v_i),
\end{equation}
where $\mathcal{O}(j)=\{i\mid (i,j)\in\mathcal{M}\}$. \clearpage


In our experiments \method\ applies \emph{all-vs-all} matchup policy, i.e., every discriminator is evaluated against every generator in the population at each generation, providing comprehensive competitive feedback across populations.

\subsection{Variation operator: training as mutation}
\label{subsec:method-variation}

The variation operator of the co-evolutionary loop is defined as a \emph{mutation} of the network parameters induced by stochastic gradient descent.
Specifically, for a given matchup $(i,j)$, we apply a number $n_t$ of training epochs of Algorithm~\ref{alg:sslgan-training} to the coupled SSL-GAN $(G_{u_j},D_{v_i})$ using mini-batches from $p_{la}$ and $p_{un}$.
This produces updated parameters $u_j'$ and $v_i'$ and therefore new individuals $G_{u_j'}$ and $D_{v_i'}$.

\subsection{Selection and elitism}
\label{subsec:method-selection}

After variation and evaluation, survival selection is applied independently in each population. The next generation is formed from $\mu$ individuals from the parent$+$offspring union.

Discriminators are ranked under Pareto dominance using the bi-objective vector
$\left(\widehat{\mathcal{L}}_{D,s}(v),\widehat{\mathcal{L}}_{D,u}(v)\right)$. 
We select the best $\mu$ discriminators by nondominated sorting and break ties with a diversity criterion (e.g., crowding distance) to preserve a well-spread approximation of the trade-off surface. 
Following the same selection method than NSGA-II (Non-dominated Sorting Genetic Algorithm II)~\cite{deb2002fast}. In contrast, generators are selected by minimizing the aggregated loss $\widehat{\mathcal{L}}_{G}(u)$.

We also consider an elitist replacement mechanism in which the best individuals in each population are preserved unchanged when forming the next generation. Elitism reduces the risk of losing high-quality solutions due to stochastic training noise and helps stabilize the co-evolutionary dynamics.

\subsection{Algorithm}
\label{subsec:method-algorithm}

Algorithm~\ref{alg:mo-sslgan} summarizes the main steps of the proposed method. The algorithm starts by initializing populations $\mathbf{G}$ and $\mathbf{D}$ (Line~\ref{alg:init_pops}). Then, individuals are evaluated using a matchup policy $\mathcal{M}$ and objective aggregation (Line~\ref{alg:evaluate_pops}). The main loop runs for $T$ generations. In each generation, offspring are created by copying selected individuals and applying SGD-based mutation by training coupled SSL-GANs for $n_t$ epochs (Line~\ref{alg:train_pairs}). After training, the union of parents and offspring is evaluated again and the next populations are obtained by survival selection (Lines~\ref{alg:select_g}--\ref{alg:select_d}). Finally, the method returns the best generator and a representative discriminator from the final non-dominated set.

\setlength{\textfloatsep}{4pt}
\setlength{\floatsep}{4pt}
\begin{algorithm}[!h]
\footnotesize
\setlength{\itemsep}{0pt}
\caption{Population-based multi-objective SSL-GAN training}
\label{alg:mo-sslgan}
\begin{algorithmic}[1]
\Require
$T$: generations,
$p_{la}$: labeled samples,
$p_{un}$: unlabeled samples,
$B_s$: batch size,
$\mu$: population size,
$n_t$: training epochs per matchup,
$\mathcal{M}$: matchup policy (all-vs-all or diagonal)
\State \textbf{Initialize} populations $\mathbf{G}=\{G_{u_1},\ldots,G_{u_\mu}\}$ and $\mathbf{D}=\{D_{v_1},\ldots,D_{v_\mu}\}$ \label{alg:init_pops}
\State \textbf{Evaluate} $\mathbf{G},\mathbf{D}$ under $\mathcal{M}$ to compute $\widehat{\mathcal{L}}_{G}$ and $(\widehat{\mathcal{L}}_{D,s},\widehat{\mathcal{L}}_{D,u})$ \label{alg:evaluate_pops}
\For{$t=1$ to $T$} \Comment{Main co-evolutionary loop}
    \State \textbf{Create offspring} $\mathbf{G}^*\gets$ copy($\mathbf{G}$), $\mathbf{D}^*\gets$ copy($\mathbf{D}$)
    \ForAll{$(i,j)\in \mathcal{M}$} \label{alg:train_pairs}
        \State Train coupled pair $(G_{u_j^*},D_{v_i^*})$ for $n_t$ epochs using Algorithm~\ref{alg:sslgan-training}
    \EndFor
    \State $\mathbf{G}\gets \mathbf{G}\cup \mathbf{G}^*$,\quad $\mathbf{D}\gets \mathbf{D}\cup \mathbf{D}^*$ \Comment{Parent+offspring union}
    \State \textbf{Evaluate} $\mathbf{G},\mathbf{D}$ under $\mathcal{M}$ \label{alg:reevaluate}
    \State $\mathbf{G}\gets$ selectBest$_\mu(\mathbf{G}, \widehat{\mathcal{L}}_{G})$ \label{alg:select_g}
    \State $\mathbf{D}\gets$ selectBest$_\mu(\mathbf{D}, (\widehat{\mathcal{L}}_{D,s},\widehat{\mathcal{L}}_{D,u}))$ \label{alg:select_d}
\EndFor
\State \Return best generator $G_u\in\mathbf{G}$ and discriminator $D_v\in\mathbf{D}$ according to validation criteria 
\end{algorithmic}
\end{algorithm}

\section{Experimental setup}
\label{sec:experimental-setup}

We perform an empirical analysis of the proposed \hbox{\method} by examining both (i) the discriminator classification accuracy and (ii) the quality of the samples produced by the generator.

\begin{figure}[!h]
\centering
    \includegraphics[width=0.5\linewidth]{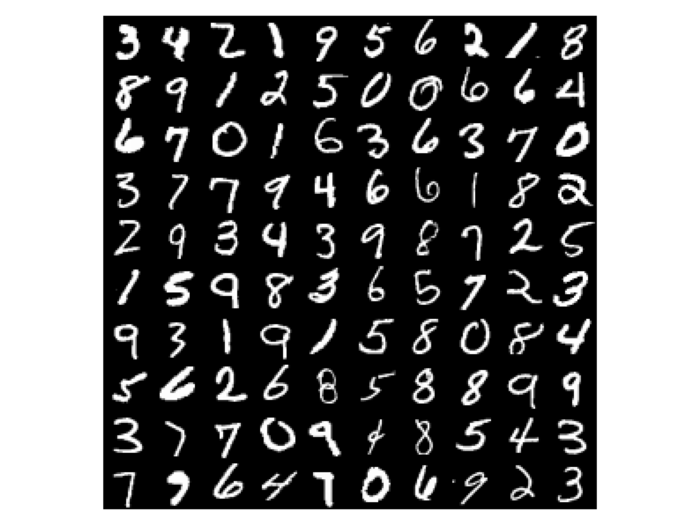}
\caption{Samples of the MNIST dataset used in the experiments.}
\label{fig:mnist-dataset}
\end{figure}

Experiments are conducted on the MNIST dataset, which contains gray-scale images of handwritten digits (0–9) with a resolution of $28\times28$ pixels (see Fig.~\ref{fig:mnist-dataset}). The dataset includes 60,000 training images and 10,000 test images. We consider a limited-label setting with 100 labeled samples per class are selected from the training set, while the remaining 59,000 images are treated as unlabeled data. Discriminator classification performance is evaluated on the MNIST test set.


The discriminator is assessed using the classification accuracy on the test set. The generator is evaluated using the Structural Similarity Index Measure (SSIM). Given two images $x$ and $y$, SSIM is computed as:
\begin{equation}
\mathrm{SSIM}(x,y) =
\frac{(2\mu_x\mu_y + c_1)(2\sigma_{xy} + c_2)}
     {(\mu_x^2 + \mu_y^2 + c_1)(\sigma_x^2 + \sigma_y^2 + c_2)},
\label{eq:ssim}
\end{equation}
where $\mu_x$ and $\mu_y$ are mean intensities, $\sigma_x^2$ and $\sigma_y^2$ are variances, and $\sigma_{xy}$ is the covariance between $x$ and $y$. Constants $c_1$ and $c_2$ stabilize the division. The resultant SSIM index is a decimal value between -1 and 1, where 1 indicates perfect similarity, 0 indicates no similarity, and -1 indicates perfect anti-correlation.

\begin{table*}[!t]
\renewcommand{\arraystretch}{0.9}
\centering
\caption{Classification accuracy results (higher is better $\uparrow$). Absolute ($\Delta$Median) and relative (Improvement (\%)) median improvements are computed with respect to the standard SSL-GAN. Best values in bold.}
\label{tab:mo_boxplot_stats_improved}
\begin{tabular}{@{}llrrrrrrrr@{}}
\toprule
Population size & Variant & Median & Q1 & Q3 & IQR & Min & Max & $\Delta$Median & Improvement (\%) \\
\midrule
&\textbf{SSL-GAN} & 0.75 & 0.67 & 0.83 & 0.16 & 0.52 & 0.90 & - & - \\
\midrule
$\mu=1$ & \textbf{base} & 0.82 & 0.82 & 0.83 & 0.01 & 0.82 & 0.83 & +0.07 & +9.33 \\
\midrule
$\mu=5$ & \textbf{base} & 0.85 & 0.81 & 0.89 & 0.08 & 0.72 & 0.92 & +0.10 & +13.34 \\
$\mu=5$ & CE-SSL-GAN & 0.85 & 0.82 & 0.88 & 0.06 & 0.74 & 0.92 & +0.10 & +13.33 \\
$\mu=5$ & \textbf{elitist}& \textbf{0.90} & 0.88 & 0.91 & 0.03 & 0.83 & 0.92 & +0.15 & +20.00 \\
\midrule
$\mu=7$ &\textbf{base}& 0.81 & 0.78 & 0.87 & 0.09 & 0.71 & 0.91 & +0.06 & +8.01 \\
$\mu=7$ & CE-SSL-GAN  & 0.82 & 0.79 & 0.87 & 0.08 & 0.70 & 0.91 & +0.07 & +9.33 \\
$\mu=7$ &  \textbf{elitist}  & \textbf{0.90} & 0.88 & 0.92 & 0.04 & 0.82 & 0.93 & +0.16 & +20.01 \\
\midrule
$\mu=9$ & \textbf{base}& 0.81 & 0.79 & 0.86 & 0.07 & 0.68 & 0.91 & +0.06 & +8.02 \\
$\mu=9$ & CE-SSL-GAN  & 0.81 & 0.78 & 0.86 & 0.08 & 0.71 & 0.91 & +0.06 & +8.00 \\
$\mu=9$ & \textbf{elitist} & \textbf{0.89} & 0.87 & 0.91 & 0.04 & 0.72 & 0.92 & +0.14 & +18.67 \\
\bottomrule
\end{tabular}
\vspace{-0.2cm}
\end{table*}

The SSL-GANs in our experiments are based on convolutional neural networks.
The generator begins with a fully connected layer mapping the latent vector into a 7$\times$7 feature map, which is then upsampled through two transposed convolutional layers with ReLU activations and batch normalization, producing a final 28$\times$28 grayscale image. The discriminator consists of four convolutional blocks that downsample the input, followed by a sigmoid-based binary \hbox{classifier}.

Our experiments compare a standard SSL-GAN training against population-based training configurations. The baseline SSL-GAN uses a single generator and discriminator trained by alternating SGD updates. The proposed approach is evaluated in three variants: (i) \textbf{base} \method, where discriminators are selected by Pareto dominance on $(\widehat{\mathcal{L}}_{D,s},\widehat{\mathcal{L}}_{D,u})$ and generators are selected by minimizing $\widehat{\mathcal{L}}_{G}$; (ii) an \textbf{elitist} \method\ variant, which preserves the best individuals unchanged when forming the next generation; and (iii) a {mono-objective} variant, i.e., CE-SSLGAN~\cite{sedeno2025}, where discriminator selection is performed using the scalarized aggregated objective.

Based on previous results in the literature~\cite{sedeno2025,casas2025}, 
we evaluate population sizes $\mu \in \{5,7,9\}$ and set the offspring size equal to the population size, i.e., $\lambda=\mu$, so that each generation produces $\mu$ offspring per population and replacement is performed on the parent+offspring union. 
The baseline SSL-GAN is run for 100 training epochs. Population-based configurations are run for 50 epochs due to their higher computational cost.
Each experiment variant is executed 30~independent times. 

The proposed methods were developed in Python. The most significant libraries used are Pytorch, Numpy, and DEAP~\cite{DEAP_JMLR2012}. The experiments were performed on a cluster of  126$\times$SD530 nodes, with 52 cores (Intel Xeon Gold 6230R @ 2.10GHz) and 192 GB of RAM.

\section{Results and Discussion}
\label{sec:results}

In this section, we present the experimental results obtained on the MNIST dataset, where 30 independent runs were performed. 
We first analyze the classification performance of the discriminator in terms of accuracy. 
Then, we assess the quality of the generated samples produced by the generator. 

The experiments include \textbf{base} \method\ with \hbox{$\mu=1$} in order to compare against standard SSL-GAN. 
When \hbox{$\mu=1$}, population effects vanish and the algorithm reduces to a single generator-discriminator pair. 
This setting allows us to isolate the impact of the Pareto-based selection mechanism used to retain the networks.

\subsection{Discriminator accuracy results}
\label{sec:disc-results}

Table~\ref{tab:mo_boxplot_stats_improved} summarizes the classification accuracy achieved by the discriminator across all evaluated configurations. 
For each method, the table reports the median accuracy over 30 independent runs together with the first and third quartiles ($Q_1$, $Q_3$), the interquartile range (IQR), and the minimum and maximum values. 
We also report the absolute improvement in median accuracy ($\Delta$Median) with respect to the standard SSL-GAN baseline, and the relative improvement computed as
$\text{Improvement}(\%) = \frac{\Delta\text{Median}}{\text{Median}_{SSL\text{-}GAN}}\times 100$.


\setlength{\textfloatsep}{4pt}
\setlength{\floatsep}{4pt}
\begin{figure*}[t]
\centering

\begin{subfigure}{0.3\textwidth}
    \centering
    \includegraphics[width=\linewidth]{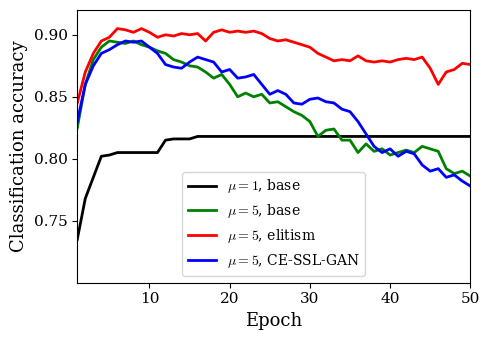}
    \caption{$\mu=1$ and $\mu=5$}
    \label{fig:pop5_evolution}
\end{subfigure}
\hfill
\begin{subfigure}{0.3\textwidth}
    \centering
    \includegraphics[width=\linewidth]{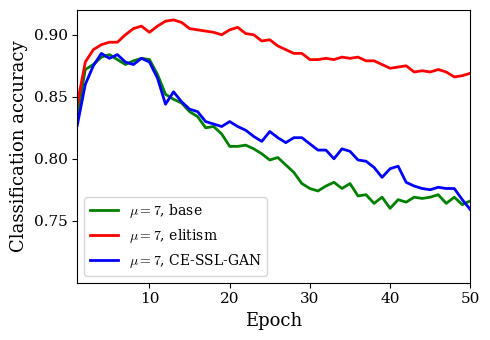}
    \caption{$\mu=7$}
    \label{fig:pop7_evolution}
\end{subfigure}
\hfill
\begin{subfigure}{0.3\textwidth}
    \centering
    \includegraphics[width=\linewidth]{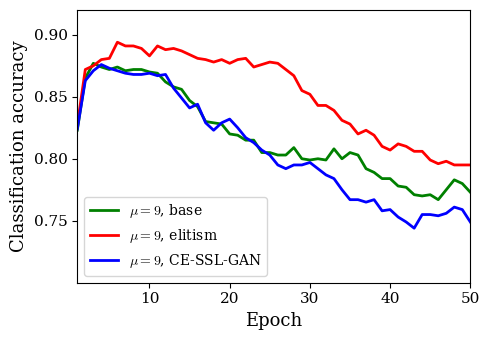}
    \caption{$\mu=9$}
    \label{fig:pop9_evolution}
\end{subfigure}

\caption{Evolution of the discriminator classification accuracy during training for different population sizes.}
\label{fig:population_evolution}
\vspace{-0.3cm}
\end{figure*}

When comparing the standard SSL-GAN with the \textbf{base} configuration of \method\ using $\mu=1$, a clear improvement is observed.
The median accuracy increases from $0.75$ to $0.82$, corresponding to a relative improvement of $+9.33\%$.
In addition, the variability across runs is substantially reduced, as reflected by the decrease of the IQR from $0.16$ to $0.01$.
According to the Wilcoxon statistical test with $\alpha<0.01$, this improvement is statistically significant.
These results indicate that the Pareto-based selection mechanism introduced in \method\ already improves discriminator training even when population effects are removed.

When population-based configurations are considered \hbox{($\mu>1$)}, the results further improve.
Across all tested population sizes, the median accuracy ranges between $0.81$ and $0.90$, corresponding to relative improvements between approximately $8\%$ and $20\%$ with respect to the SSL-GAN baseline.
This indicates that incorporating multiple interacting generators and discriminators helps the training process discover more effective discriminator strategies.
However, increasing the population size does not always lead to additional gains.
For example, the \textbf{base} configuration achieves its best median accuracy with $\mu=5$ ($0.85$), while larger populations ($\mu=7$ and $\mu=9$) produce slightly lower median values ($0.81$).
These observations are consistent with previous findings in the literature~\cite{sedeno2025,casas2025}, where increasing the population size beyond a certain point does not necessarily improve performance.
This suggests that population-based training mainly benefits from maintaining diversity rather than simply increasing the number of individuals.
Indeed, when comparing the same variant across different population sizes, the differences are not statistically significant.

Finally, when comparing the different population-based variants, the \textbf{elitist} configuration consistently achieves the best performance.
For $\mu=5$ and $\mu=7$, it reaches a median accuracy of $0.90$, corresponding to an absolute improvement of $+0.15$ and a relative improvement of $+20\%$ over SSL-GAN.
Moreover, the elitist variant produces more robust results, as indicated by its smaller IQR values (between $0.03$ and $0.04$).
Wilcoxon tests with Bonferroni correction confirm that the elitist variant significantly outperforms the other approaches.
A possible explanation is that elitism preserves the best-performing discriminator solutions across generations, preventing high-quality classifiers from being lost due to stochastic updates during training.
This stabilizes the co-evolutionary dynamics and allows the population to progressively refine strong discriminator strategies.

Regarding the evolution of the discriminator accuracy during training, Fig.~\ref{fig:population_evolution} shows the median classification accuracy across training epochs for the analyzed variants in a randomly chosen representative run.

For population sizes greater than one, a consistent pattern appears across methods. During the initial training phase, discriminator accuracy increases rapidly, reaching its highest values around epochs 10-20. After this stage, the accuracy gradually declines. This behavior is observed across population sizes and variants, although its magnitude depends on the training strategy.

In contrast, when the population size is $\mu=1$, accuracy improves more gradually and then stabilizes. This reflects the absence of competitive interactions among individuals, resulting in a more stable but less exploratory training process. In some runs, the final accuracy obtained with $\mu=1$ even surpasses the values achieved by the \textbf{base} and CE-SSL-GAN variants, indicating that population-based methods may experience performance degradation during training.

The \textbf{elitist} variant consistently maintains the highest accuracy throughout training. By preserving the best-performing individuals, elitism reduces the impact of stochastic updates and prevents strong discriminators from being replaced by weaker offspring. Consequently, the elitist configuration exhibits a slower and less pronounced degradation compared with the other variants.

The degradation in discriminator accuracy becomes more pronounced as the population size increases. Larger populations introduce stronger competitive pressure between generators and discriminators, accelerating changes in the adversarial dynamics. Nevertheless, the elitist strategy mitigates this effect by preserving the best classifiers across generations.

\begin{figure}[!h]
\centering
    \includegraphics[width=0.5\linewidth]{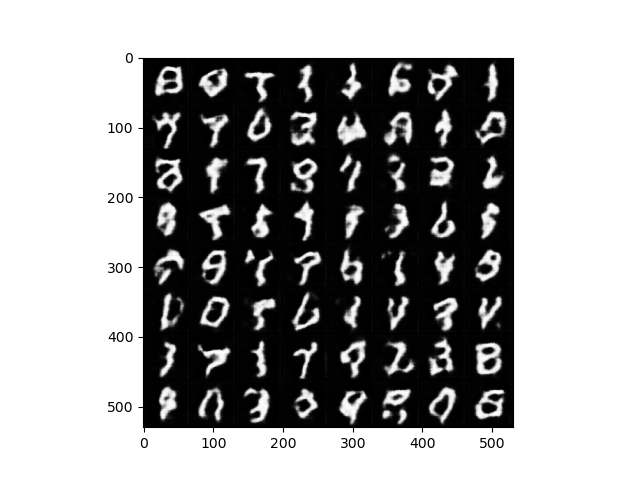}
\caption{Generated samples by the \textbf{base} \method\ with $\mu=5$ at training epoch = 10.}
\label{fig:bad-samples}
\end{figure}

Finally, the decrease in discriminator accuracy can be explained by the improving quality of generated samples. As training progresses, the generator produces samples increasingly similar to real data, making the discriminative task more difficult~\cite{dai2017good}. Fig.~\ref{fig:bad-samples} shows samples generated by the \textbf{base} variant with $\mu=5$ at epoch 10, when the generator still produces low-quality samples that are easier for the discriminator to distinguish. As the generator improves in later epochs (see Fig~\ref{fig:ssim_samples_comparison}), the task becomes harder, contributing to the observed reduction in classification accuracy.

\begin{table*}[!t]
\renewcommand{\arraystretch}{0.9}
\centering
\caption{SSIM results (higher is better $\uparrow$). Absolute ($\Delta$Median) and relative (Improvement (\%)) median improvements are computed with respect to the standard SSL-GAN.}
\label{tab:ssim_boxplot_stats_improved}
\begin{tabular}{@{}llrrrrrrrr@{}}
\toprule
Population size & Variant & Median & Q1 & Q3 & IQR & Min & Max & $\Delta$Median & Improvement (\%) \\
\midrule
& \textbf{SSL-GAN} & 0.32 & 0.26 & 0.37 & 0.11 & 0.16 & 0.45 & - & - \\
\midrule
$\mu=1$ & \textbf{base}  & 0.39 & 0.38 & 0.39 & 0.01 & 0.38 & 0.39 & +0.07 & +20.76 \\
\midrule
$\mu=5$ & \textbf{base}  & 0.38 & 0.37 & 0.39 & 0.02 & 0.34 & 0.42 & +0.06 & +18.88 \\
$\mu=5$ & CE-SSL-GAN & 0.39 & 0.36 & 0.40 & 0.04 & 0.34 & 0.45 & +0.07 & +21.67 \\
$\mu=5$ & \textbf{elitist} & 0.38 & 0.34 & 0.39 & 0.04 & 0.32 & 0.44 & +0.06 & +17.97 \\
\midrule
$\mu=7$ & \textbf{base} & 0.41 & 0.39 & 0.42 & 0.03 & 0.36 & 0.45 & +0.09 & +27.41 \\
$\mu=7$ & CE-SSL-GAN & 0.41 & 0.40 & 0.42 & 0.02 & 0.38 & 0.44 & +0.09 & +28.66 \\
$\mu=7$ & \textbf{elitist} & 0.38 & 0.37 & 0.40 & 0.03 & 0.35 & 0.43 & +0.06 & +19.57 \\
\midrule
$\mu=9$ & \textbf{base}  & 0.40 & 0.39 & 0.41 & 0.02 & 0.38 & 0.43 & +0.08 & +26.56 \\
$\mu=9$ & CE-SSL-GAN & 0.41 & 0.40 & 0.42 & 0.03 & 0.38 & 0.44 & +0.09 & +28.91 \\
$\mu=9$ & \textbf{elitist}  & 0.41 & 0.40 & 0.42 & 0.03 & 0.37 & 0.44 & +0.09 & +27.88 \\
\bottomrule
\end{tabular}
\end{table*}

\subsection{Generated sample quality (SSIM) results}
\label{sec:ssim-results}

\begin{figure}[h!]
\vspace{-0.1cm}
    \centering
    
    \begin{subfigure}[t]{0.3\linewidth}
        \centering
        \includegraphics[width=\linewidth,trim={3.2cm 0cm 3.2cm 0cm}, clip]{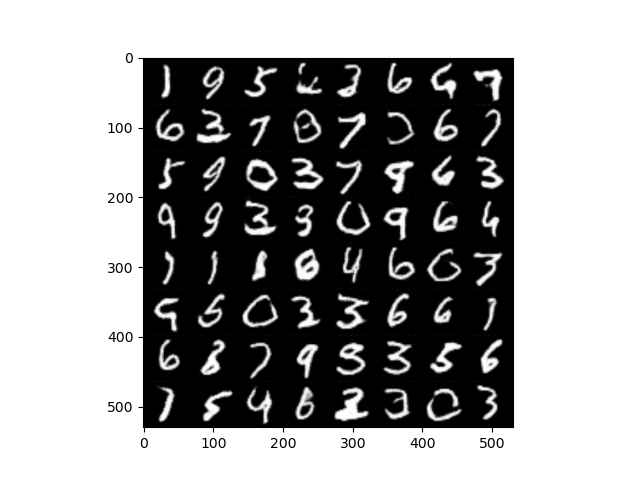}
        \caption{\textbf{base}}
        \label{fig:mo_ssl_gan}
    \end{subfigure}
\hfill
    \begin{subfigure}[t]{0.3\linewidth}
        \centering
        \includegraphics[width=\linewidth,trim={3.2cm 0cm 3.2cm 0cm}, clip]{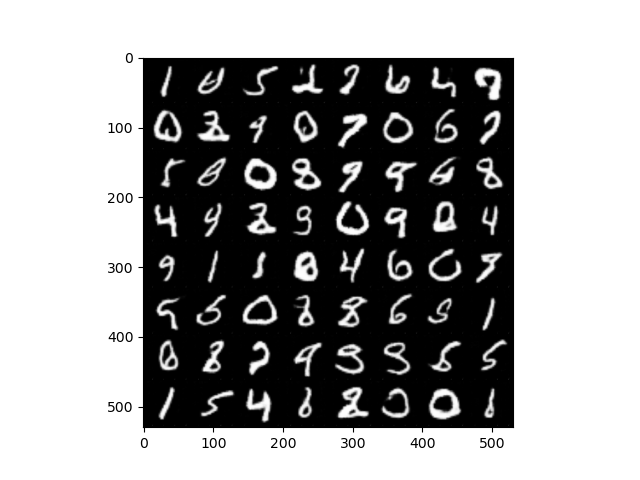}
        \caption{CE-SSL-GAN}
        \label{fig:ce_ssl_gan}
    \end{subfigure}
    \hfill
    \begin{subfigure}[t]{0.3\linewidth}
        \centering
        \includegraphics[width=\linewidth,trim={3.2cm 0cm 3.2cm 0cm}, clip]{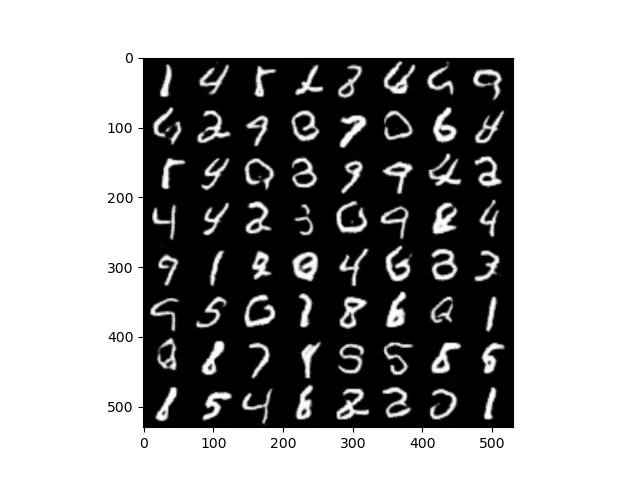}
        \caption{\textbf{elitist}}
        \label{fig:elitist}
    \end{subfigure}
    
    \caption{Generated samples by the analyzed  coevolutionary SSL-GAN variants with $\mu=5$.}
    \label{fig:ssim_samples_comparison}
    \vspace{-0.1cm}
\end{figure}

Table~\ref{tab:ssim_boxplot_stats_improved} summarizes the SSIM obtained for the generated samples across all evaluated configurations.
For each method, the table reports the median SSIM over 30 independent runs together with the first and third quartiles ($Q_1$, $Q_3$), the interquartile range (IQR), and the minimum and maximum values.
We also report the absolute and relative improvement in median SSIM with respect to the standard SSL-GAN baseline.
In addition to the quantitative analysis, Fig.~\ref{fig:ssim_samples_comparison} illustrates representative samples generated by the coevolutionary variants of SSL-GAN with $\mu=5$.

When comparing the standard SSL-GAN with the \textbf{base} configuration of \method\ using $\mu=1$, a clear improvement in the quality of generated samples is observed.
The median SSIM increases from $0.32$ to $0.39$, corresponding to an absolute improvement of $+0.07$ and a relative improvement of $+20.76\%$.
Additionally, the variability across runs is substantially reduced, as indicated by the dramatic reduction in IQR from $0.11$ to $0.01$.
These results show that even without population effects, the Pareto-based selection mechanism introduced by \method\ improves the quality and stability of generated samples.

When population-based configurations are considered \hbox{($\mu>1$)}, the SSIM values improve further.
Across all population-based variants, the median SSIM ranges between $0.38$ and $0.41$, corresponding to relative improvements between approximately $18\%$ and $29\%$ with respect to the SSL-GAN baseline.
The best performance is achieved with $\mu=9$ using the CE-SSL-GAN variant, reaching a median SSIM of $0.41$.
However, despite these quantitative differences, the visual inspection of the generated digits in Fig.~\ref{fig:ssim_samples_comparison} shows that all coevolutionary variants produce samples with very similar perceptual quality.

Interestingly, increasing the population size does not always lead to consistent improvements across variants.
For example, the \textbf{base} configuration improves from $0.38$ at $\mu=5$ to $0.41$ at $\mu=7$, but slightly decreases to $0.40$ at $\mu=9$.
Similar trends appear in other variants.
This suggests that population-based training mainly benefits from maintaining diversity rather than simply increasing the number of individuals.

Finally, when comparing the different population-based variants, the CE-SSL-GAN and elitist configurations tend to produce the highest SSIM values.
The elitist variants show slightly more stable behavior across runs, as indicated by their moderate IQR values.
However, statistical tests do not reveal significant differences among the analyzed population-based variants.
Overall, these results indicate that coevolutionary training consistently improves SSIM with respect to standard SSL-GAN, while producing visually comparable samples across the different evolutionary configurations.

\subsection{General discussion}

The results highlight the impact of the Pareto-based discriminator selection introduced in \method. 
Even without population effects ($\mu=1$), the method improves both discriminator accuracy and SSIM compared with standard SSL-GAN. 
This suggests that treating discriminator training as a multi-objective optimization problem enables a better balance between supervised classification and unsupervised real/fake discrimination. 
In contrast to approaches such as CE-SSL-GAN, which aggregate these objectives into a single scalar loss~\cite{sedeno2025}, the Pareto-based formulation explicitly preserves the trade-off between them. 
This allows the discriminator to explore multiple solutions along the objective frontier rather than collapsing the objectives into a single optimization target.

When population-based configurations are considered, the interaction between multiple generators and discriminators further improves training performance. 
However, increasing the population size does not consistently lead to additional gains, indicating that the main benefit of population-based training arises from maintaining diversity rather than simply increasing the number of individuals.

Among the analyzed variants, the \textbf{elitist} configuration consistently achieves the best discriminator accuracy and the most stable results across runs. 
By preserving the best individuals across generations, elitism prevents strong discriminators from being replaced by weaker offspring and stabilizes the coevolutionary dynamics.

The experiments reveal a clear interaction between generator improvement and discriminator accuracy. 
As the generator produces more realistic samples, the classification task becomes harder for the discriminator, leading to the decrease in accuracy observed during later training stages~\cite{dai2017good}.

Therefore, the results suggest that combining multi-objective discriminator training with competitive coevolution provides a robust framework for stabilizing SSL-GAN training, improving discriminator classification performance, and producing higher-quality generated samples. \\

\section{Conclusions and Future Work}
\label{sec:conclusions}

This paper investigated \method, a population-based training strategy for SSL-GANs in which discriminator learning is formulated as a multi-objective optimization problem. Instead of aggregating the supervised and unsupervised components of the SSL objective into a single loss, the proposed approach maintains a population of discriminators evaluated through Pareto dominance, allowing the training process to explore different trade-offs between classification accuracy and real/fake discrimination.

The experimental results on MNIST with limited labeled data show that the proposed training framework improves the robustness of SSL-GAN optimization. Even without population effects, the Pareto-based selection mechanism produces more stable and accurate discriminators than the standard SSL-GAN baseline and CE-SSL-GAN. When population-based configurations are considered, the interaction between multiple generators and discriminators further improves training dynamics. In particular, the \textbf{elitist} configuration consistently achieves the best classification accuracy and the most stable results across runs. The results also indicate that population-based training improves the consistency of generated samples, although the differences between evolutionary variants remain moderate. Therefore, these findings suggest that combining multi-objective optimization with competitive co-evolution, extending ideas explored in CE-SSL-GAN, provides a promising direction for stabilizing SSL-GAN training.

Future work includes:
(i) evaluating the approach on more complex datasets such as CIFAR-10 and SVHN;
(ii) defining generator training as a multi-objective problem, considering both fidelity and diversity objectives rather than only fidelity;
(iii) exploring alternative multi-objective selection strategies; and
(iv) studying distributed implementations to scale the population-based training process.

\section*{Acknowledgment}
This research is partially funded by the Universidad de Málaga (UMA) under grant PID2024-158752OB-I00 (AIM-ZERO) by MICIU/AEI/10.13039/501100011033. The authors thank the Supercomputing and Bioinformatics center at the UMA for their computer resources and assistance. Funding for open access charge: Universidad de Málaga / CBUA.

\bibliographystyle{IEEEtran}

\end{document}